\documentclass{article}

\usepackage{microtype}
\usepackage{graphicx}
\usepackage{subfigure}
\usepackage{booktabs} 
\usepackage{float}

\usepackage{hyperref}

\usepackage{xurl}

\usepackage[accepted]{packages/icml2025}

\usepackage{amsmath}
\usepackage{amssymb}
\usepackage{mathtools}
\usepackage{amsthm}

\usepackage[capitalize,noabbrev]{cleveref}

\crefname{appendix}{Appendix}{Appendices}
\Crefname{appendix}{Appendix}{Appendices}

\theoremstyle{plain}

\theoremstyle{definition}

\theoremstyle{remark}

\usepackage[textsize=tiny]{todonotes}

\icmltitlerunning{Submission and Formatting Instructions for ICML 2025}

\begin{document}

\twocolumn[
  \icmltitle{Access Controls Will Solve the Dual-Use Dilemma}



  \icmlsetsymbol{equal}{*}

  \begin{icmlauthorlist}
    \icmlauthor{Evžen Wybitul}{eth}
  \end{icmlauthorlist}

  \icmlaffiliation{eth}{ETH Zurich, Switzerland}

  \icmlcorrespondingauthor{Evžen Wybitul}{wybitul.evzen@gmail.com}

  \icmlkeywords{Gradient Routing, Modularization, AI Safety,
  Unlearning, Access Control, Technical AI Governance}

  \vskip 0.3in
]



\printAffiliationsAndNotice{} 

\begin{abstract}
  AI safety systems face the dual-use dilemma. It is unclear whether to answer dual-use requests, since the same query could be either harmless or harmful depending on who made it and why. To make better decisions, such systems would need to examine requests' real-world context, but currently, they lack access to this information. Instead, they sometimes end up making arbitrary choices that result in refusing legitimate queries and allowing harmful ones, which hurts both utility and safety. To address this, we propose a conceptual framework based on access controls where only verified users can access dual-use outputs. We describe the framework's components, analyse its feasibility, and explain how it addresses both over-refusals and under-refusals. While only a high-level proposal, our work takes the first step toward giving model providers more granular tools for managing dual-use content. Such tools would enable users to access more capabilities without sacrificing safety, and offer regulators new options for targeted policies.
\end{abstract}

\section{Introduction} \label{section:introduction}

\emph{What features of viral surface proteins are recognized by human
antibodies?}

\begin{figure}[t]
  \vskip 0.2in
  \begin{center}
    \centerline{\includegraphics[width=\columnwidth]{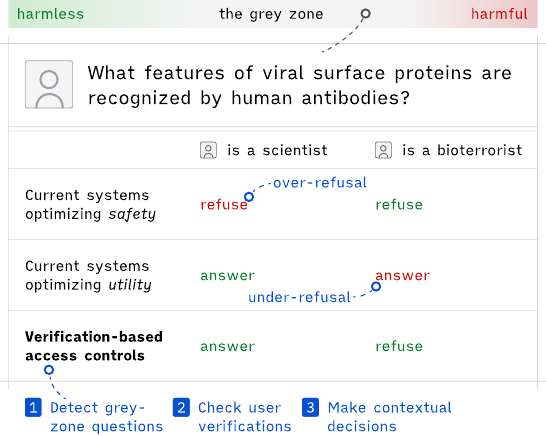}}
    \caption{
      The dual-use dilemma: the same question could be harmless or harmful depending on who asks it.
      Should we refuse to answer it?
      Current safety systems must decide without knowing the user's context.
      This leads to over-refusals (blocking legitimate users) and under-refusals (allowing bad actors).
      Verification-based access controls solve this by detecting grey-zone questions, obtaining real-world context about users, and making contextual decisions that can get both cases right.
    }
    \label{figure:main}
  \end{center}
  \vskip -0.2in
\end{figure}

Is this question safe to answer?
While some user requests and large language model outputs are clearly benign or harmful, many fall in the grey zone in the middle, including the question above.
In the grey zone, the harmfulness of a request depends not on its content, but on its \emph{real-world context}: who made it and for what purpose.
We illustrate this in \cref{figure:main}.

Safety systems that rely solely on content analysis immediately face the \emph{dual-use dilemma}.
When confronted with a grey-zone request, should they refuse it or not?
This forces arbitrary decisions that reduce both utility and safety: some legitimate queries are refused (over-refusals) while some harmful ones are not (under-refusals).
Certain safety systems attempt to address this by inferring the context of the request from its contents or the chat history.
However, this inferred context is based entirely on user-provided inputs and can be easily fabricated by adversaries.

In this paper, we argue that informative, hard-to-fabricate real-world context can be obtained through user-level verifications such as ID checks, institutional affiliations, or government-issued certifications.
We address the dual-use dilemma with two contributions.
\textbf{As our primary contribution}, we show how this verified context can be used jointly with content analysis in a safety framework based on access controls \cite{butler1974}.
In the framework, model outputs are classified into content categories, and the system verifies whether the user has the required credentials to access the detected category.
We also describe how the framework addresses both issues caused by the dual-use dilemma: over-refusals and under-refusals.
\textbf{Additionally}, we propose a novel theoretical approach to content category classification based on recent methods for robust unlearning, UNDO \cite{lee2025distillationrobustifiesunlearning} and gradient routing \cite{cloud2024gradientroutingmaskinggradients}.
This approach avoids the capability gap between a model and its monitors that can make output monitoring methods non-robust \cite{jin2024jailbreakinglargelanguagemodels}.

Current approaches to AI safety force crude trade-offs between blanket restrictions that stifle beneficial uses and permissive policies that enable misuse.
Our framework represents a first step toward solving the challenge of ``detection and authorization of dual-use capability at inference time'' highlighted by a recent survey of problems in technical AI governance \cite{reuel2025openproblemstechnicalai}.
By giving model providers better tools for contextual safety decisions, access control frameworks could benefit all stakeholders --- enabling users to access more capabilities whilst providing regulators with a way to avoid blunt regulatory instruments.

\section{Current Safety Methods Don't Solve the Dual-Use Dilemma}
\label{section:current-methods}

The dual-use dilemma causes two issues: over-refusals and under-refusals.
Over-refusals reduce model utility for legitimate users, which is clearly undesirable.
Under-refusals are equally problematic because they enable decomposition attacks \cite{glukhov2023llmcensorshipmachinelearning, glukhov2024breachthousandleaksunsafe}.
These attacks transform clearly harmful queries, such as ``How to modify a virus to avoid immune detection?'', into series of mundane grey-zone questions, such as the question about viral surface proteins from \cref{figure:main}.
Safety systems would refuse the harmful query but do not refuse the grey-zone ones.
Through these attacks, adversaries exploit under-refusals in a way that compromises system safety.

Since whether a grey-zone request should be refused depends on who made it and why, preventing both over-refusals and under-refusals requires access to real-world context.
This means that the traditional focus on resilience against jailbreak and prompt injections is orthogonal to this problem.
Instead, we evaluate three approaches from the AI safety literature to see how sensitive they are to contextual information, and whether their sources of real-world context are trustworthy --- that is, hard to manipulate.

\subsection{Unlearning: Non-Contextual Removal of Concepts}

Unlearning methods aim to remove specific knowledge, concepts, or capabilities from a model after training \cite{liu2024rethinkingmachineunlearninglarge}.
Their goal is to eliminate the model's ability to generate harmful content while preserving other capabilities.

Unlearning faces significant technical challenges even for preventing behaviours that are clearly harmful.
As noted by \citet{cooper2024machineunlearningdoesntthink} and \citet{barez2025openproblemsmachineunlearning}, capabilities are hard to define, hard to remove without side effects, and hard to trace back to specific data points.
Furthermore, many unlearning approaches mask rather than truly remove the targeted knowledge \cite{deeb2025unlearningmethodsremoveinformation}.

\subsection{Safety Training: The Model Reacts to Context}

Safety training methods modify the model's training process to align
its outputs with human preferences.
This category includes safety pre-training \cite{maini2025safetypretraininggenerationsafe}, RLHF \cite{christiano2023deepreinforcementlearninghuman}, and safety finetuning.

Unlike unlearning, these methods are contextual.
They don't remove capabilities entirely but train the model to selectively deploy them based on, among other things, the perceived legitimacy and harmlessness of the request.
However, these qualities are entirely inferred from user-supplied information, such as the request text or chat history.
Without access to trustworthy real-world context, the model cannot make truly informed decisions about grey-zone scenarios.
For example, models are susceptible to attacks that fabricate in-chat context \cite{zeng2024johnnypersuadellmsjailbreak}, or attacks that diminish models' sensitivity to it \cite{russinovich2025greatwritearticlethat}.
While modifying these methods to incorporate external contextual information is possible in theory, it would likely result in a more opaque and less modular system than making similar modifications to post-processing methods.

\subsection{Post-Processing: External Systems React to Context} \label{section:post-processing}

Post-processing methods are systems that classify user inputs and system outputs for the purposes of steering the underlying model, or monitoring and filtering its responses.
Sometimes, these methods are used for usage monitoring, as is the case with Anthropic's Clio \cite{tamkin2024clioprivacypreservinginsightsrealworld, handa2025economictasksperformedai}, other times, they are used for safety, as with Llama Guard \cite{inan2023llamaguardllmbasedinputoutput} and Constitutional Classifiers \cite{sharma2025constitutionalclassifiersdefendinguniversal}.
However, similarly to safety training, the ``real-world'' context these methods work with is currently inferred mostly from user-supplied content and thus untrustworthy and vulnerable to attacks, as evidenced by the many jailbreaks that successfully target current production systems \cite{zhang2025outputconstraintsattacksurface}.
Nevertheless, these methods could be modified to incorporate external contextual information, potentially serving as a foundation for more trustworthy, contextual safety mechanisms. We discuss this option in \cref{section:content-classification}.

\section{Access Controls as a Solution}
\label{section:access-controls}

Current safety systems face the dual-use dilemma because they lack trustworthy information about who is making the request and why.
In this section, we describe an access control system that addresses this problem.

\subsection{Overview of the Access Control Framework} \label{section:access-control-overview}

We propose a defensive system where grey-zone requests are refused by default, but users can gain access to specific categories of knowledge if they undergo verification.

When model providers set up the system, they will make two core design choices with the help of domain experts.
First, they will define \textbf{content categories} (\cref{section:content-categories}): groups of sensitive topics organized by domain and risk rating.
Second, for each content category, they will specify a \textbf{verification mechanism} (\cref{section:verification-mechanisms}): the verification process users must complete to access that category.

Whenever the model generates an output, the system will perform \textbf{content classification} (\cref{section:content-classification}) to check whether the model's output belongs to any of the predefined content categories.
If the user lacks authorization for the detected category, the system will take graduated \textbf{system responses} (\cref{section:system-responses}) ranging from enabling enhanced logging to a full refusal.

\begin{table}[t]
  \caption{Example content categories in pathogen biology.}
  \vspace{0.1in}
  \label{table:biology-examples}
  \centering
  \small
  \begin{tabular}{lll}
    \toprule
    \textbf{Risk Level} & \textbf{Examples} & \textbf{Verification} \\
    \midrule
    Low & Basic knowledge & --- \\
    Moderate & CRISPR protocols & ID verification \\
    High & Viral surface proteins & Biosafety certification \\
    \bottomrule
  \end{tabular}
\end{table}

While the exact categorization will be domain-specific, we anticipate a common pattern where the majority of content remains freely accessible, with progressively more stringent verification requirements for higher-risk categories.
\Cref{table:biology-examples} illustrates how this might look in pathogen biology.
Under this example, if a user asks the question about viral surface proteins, the system would detect that the request belongs to a high-risk category, check whether the user has the required biosafety certification, and either provide the information or prompt them to complete verification first.

This approach directly addresses the dual-use dilemma by preventing under-refusals and reducing over-refusals.
Decomposition attacks become much harder because the system refuses grey-zone requests by default --- attackers would need legitimate credentials rather than clever prompting.
Simultaneously, verified users gain access to specialized knowledge that would otherwise face blanket restrictions under current approaches.
We discuss the feasibility and limitations of this framework, including considerations around user friction, in \cref{section:feasibility-and-limitations}.

\subsection{Content Categories} \label{section:content-categories}

Model providers will develop content categories by adapting existing risk frameworks with the help of domain experts.
In biology, experts could build on biosafety levels (BSL)~\cite{CDC_BMBL_2020} and dual-use research of concern policies~\cite{USG_DURC_2012}.
However, since existing frameworks typically categorize only high-level concepts like organisms or compounds, experts would need to decompose them into smaller, more specific components suitable for knowledge access control.

For instance, cultivating and handling a dangerous BSL-3 pathogen might involve multiple distinct knowledge components: (1) specific procurement methods, (2) cultivation techniques, (3) purification methods, and (4) protocols for specialized equipment.
For each component, experts would assess how often it enables harmless versus harmful applications, then assign it to an appropriate risk category.
This decomposition approach transforms broad categories into granular knowledge components that can be individually controlled, as illustrated by the different risk levels in \cref{table:biology-examples}.

Evidence from chemistry suggests this approach is at least sometimes feasible: the risk schedules of the Chemical Weapons Convention already identify not just controlled compounds but also their precursors~\cite{OPCW_CWC_1993}, demonstrating successful decomposition into components.
Nevertheless, some harmful applications might not decompose so neatly; we discuss this limitation in \cref{section:feasibility-and-limitations}.

\subsection{Verification Mechanisms} \label{section:verification-mechanisms}

Each content category requires a verification process that users must complete to access it, as shown in \cref{table:biology-examples}.
Rather than creating new systems, model providers will build on existing verification infrastructure, consulting domain experts to identify appropriate mechanisms for each field.

For moderate-risk categories, model providers could use established identity verification services like Stripe Identity~\cite{stripe_identity_2024} or institutional systems like ORCID~\cite{orcid_2024}.
These systems provide global, standardized, low-friction solutions with one-time costs under \$2 per user.
They would serve primarily to maintain audit trails for post-incident investigation and provide a deterrent effect.

High-risk categories could leverage existing domain-specific certifications that demonstrate users' ability to handle sensitive information and materials responsibly.
In biology, this might include governmental certifications for handling high BSL pathogens, and equivalent certifications in other countries.

This approach faces several limitations, including the risk of being overly restrictive and concerns about equitable access across different countries.
We discuss these limitations and potential solutions in \cref{section:feasibility-and-limitations}.

\subsection{Implementing Content Classification} \label{section:content-classification}

Model providers will need to classify model outputs into content categories during generation.
We examine three possible implementations below.
While none of these approaches have been empirically validated for risk category classification specifically, each represents a viable technical path that could be developed and evaluated by practitioners interested in implementing access controls.
We leave the discussion of how classification errors might influence user experience for \cref{section:feasibility-and-limitations}.

\paragraph{Separate Models}

The most straightforward approach is to train separate models with different capabilities, and route users to the appropriate model based on their authorization.

This approach offers strong robustness against adversarial attacks since unauthorized knowledge is physically absent from the model.
However, this approach proves impractical for real deployment, as model providers would need to train and maintain potentially dozens of model variants.

\begin{figure}[t]
  \vskip 0.2in
  \begin{center}
    \centerline{\includegraphics[width=\columnwidth]{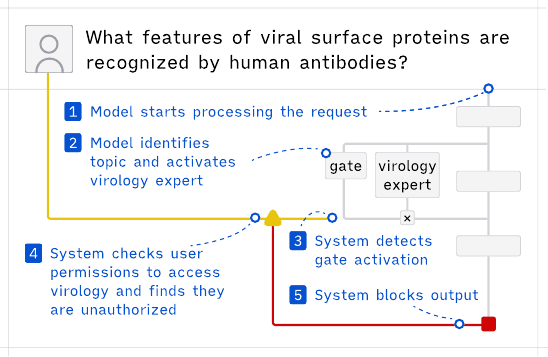}}
    \caption{
      The schema shows how an access control system could be implemented with specialized expert modules.
      (1)~The model begins to answer the question because it is trained to be helpful.
      (2)~During the forward pass, the model detects the question is about virology and activates its virology expert module that contains relevant knowledge.
      (3)~The activation of the expert is observed by an external mechanism that (4)~checks in the company's database if the user has the required authorization to access virology knowledge.
      (5)~Since they don't, the model is stopped.
      If they did, the model would be allowed to give an answer.
    }
    \label{figure:experts}
  \end{center}
  \vskip -0.2in
\end{figure}

\paragraph{Specialized Expert Modules}

Instead of maintaining separate models, model providers could use a single model with separate expert modules that activate when their specialized knowledge is required.
\Cref{figure:experts} illustrates how this approach might work when a user poses a grey-zone question.

To implement this, we could concentrate distributed knowledge into specialized expert modules using a novel method that would combine UNDO \cite{lee2025distillationrobustifiesunlearning} and gradient routing \cite{cloud2024gradientroutingmaskinggradients}.
The approach would first unlearn knowledge belonging to content categories from the base model, then distil this unlearned model into a new model with expert modules for each category.
During distillation, gradients from examples in each content category would be routed exclusively through their associated expert modules, while the model would be trained to activate experts only when generating relevant content.
Concurrent work adopted and successfully evaluated a similar approach for controlling access to information the model learned through fine-tuning \cite{jayaraman2025permissionedllmsenforcingaccess}.
Our proposed approach is more general and could be applied to knowledge obtained in any training stage.

This approach could offer several advantages.
It would likely add minimal latency since the expert modules would be small.
More importantly, it could provide strong robustness: if an attacker prevented expert module activation to avoid detection, they would simultaneously prevent access to the specialized knowledge stored in that module, making the attack self-defeating.
While this method remains entirely theoretical and requires empirical validation, these potential properties make it worth investigating.

\paragraph{Post-Processing}

Moving from theoretical approaches to more practical ones, post-processing methods such as Constitutional Classifiers \cite{sharma2025constitutionalclassifiersdefendinguniversal} offer a proven approach to content classification.
They operate independently of the model, allowing for rapid deployment and iteration, and they could be adapted to detect content categories and trigger checks of user verifications.
However, for latency reasons, there is sometimes a capability gap between the model and the post-processing system, which adversaries can exploit to evade detection \cite{jin2024jailbreakinglargelanguagemodels, kumar2025freelunchguardrails}.

\subsection{System Responses} \label{section:system-responses}

When content classification detects restricted categories, the system can implement various responses depending on the risk level and classification confidence.
This provides an additional parameter that model providers can tune based on their specific stringency requirements.

For example, outputs classified as restricted with high confidence might be immediately refused, with the system providing a message indicating which verification is required for access.
For borderline classifications where confidence is low, the system might allow response generation while enabling enhanced logging and additional safety review before serving the output to the user.
Other possible responses include content generation with a steered model, or graduated restrictions for repeat violations.

\section{Feasibility and Limitations}
\label{section:feasibility-and-limitations}

\Cref{section:access-controls} identified several technical challenges: (1)~some harmful knowledge might decompose into concepts that are all indispensable for harmless applications; (2)~some verifications may be too difficult to obtain; and (3)~content classifiers might produce false positives.
These challenges increase user friction and impose utility costs.
We analyse how access controls affect the safety-utility trade-off by examining over-refusals and under-refusals separately, as this highlights the different impacts our system has in each case.

\subsection{Access Controls Help Against Over-Refusals}

For over-refusals, access controls can provide a strict improvement over the status quo, even with the technical challenges taken into account.
Model providers can only require verification for requests that would otherwise be refused.
This is a clear improvement in utility: legitimate users can either accept refusal (current experience) or complete verification to gain access (new option).
Safety need not be hurt: model providers can design the verification requirements to be strict enough to deter the overwhelming majority of adversaries.

\subsection{Domain-specific Approaches to Under-Refusals}

Using access controls to prevent under-refusals means making some accessible knowledge require verification.
Thus, even a perfect system takes a toll on utility since legitimate users must get verified, with additional costs caused by the technical challenges mentioned above.

Model providers can measure the impacts of the system on utility before deployment, for example by conducting partial rollouts.
They can then walk the safety-utility frontier by tuning the parameters of the system: making verification more or less stringent, raising or lowering classifier thresholds, and so on.
The optimal point on the frontier will depend on (and will change over time with) the domain-specific technological constraints, liability requirements, regulatory pressures, and business priorities.
In many domains, the default position might be the status quo, which offers maximum utility.
However, this is not true for some domains like pathogen biology, in which the trade-off for safety is already favourable, as the following example demonstrates.

Models contain a lot of safety-relevant knowledge about pathogen biology; a recent study found that frontier models have more tacit knowledge than 94\% of expert virologists \cite{gotting2025virologycapabilitiestestvct}.
Consider an access control system so badly calibrated that all pathogen-related queries would require verification.
This represents an upper bound on friction since real implementations would be far more targeted.
First, we note that this system would deter at least some adversaries from performing decomposition attacks, improving on the safety of the status quo.
At the same time, the impact on utility is minimal: only 0.85\% of queries would experience added friction (estimated based on the Anthropic Economic Index, see \cref{appendix:estimating-biology-requests} for details).
This would likely be acceptable to model providers; for context, existing safety systems like Constitutional Classifiers incorrectly refuse around 0.5\% queries \cite{sharma2025constitutionalclassifiersdefendinguniversal}.

\subsection{The Incentives Grow with Model Scale}

As models gain even more sophisticated knowledge across dual-use domains, both the value of the knowledge for specialized professional users and the misuse risks it poses increase.
Thus, controlling both under-refusals and over-refusals will be more and more important, and the business case for implementing access controls strengthens.

\subsection{Limitations and Open Questions}

Since the system links real-world identities to requests, it poses risks of surveillance and information leakage.
Model providers should establish clear privacy policies and use privacy-preserving systems like Clio \cite{tamkin2024clioprivacypreservinginsightsrealworld} for logging and auditing.

Additionally, using verification mechanisms based on certifications that are not available globally could exclude legitimate researchers from access.
Model providers should work with regional authorities to address this and offer interim manual approval processes for users facing structural barriers.

\section{Conclusion}

Current safety systems face a dual-use dilemma: should they refuse a request that could be either harmless or harmful depending on who made it and why?
Our access control framework addresses this by incorporating user verification into safety decisions, reducing over-refusals for legitimate users whilst preventing under-refusals that enable decomposition attacks.
We also propose a novel content classification approach that could offer high efficiency and robustness to attacks, though this theoretical contribution requires empirical validation.
As models develop increasingly sophisticated dual-use knowledge, model providers need better tools for managing the tension between utility and safety.
Access control frameworks offer a path toward more nuanced safety decisions that could benefit all stakeholders --- improving outcomes for users whilst providing regulators with granular governance options that avoid crude policy trade-offs.

\section*{Acknowledgements}

We thank Jakub Kryś, and Dennis Akar for their feedback on a draft of this paper. We thank Joseph Miller, Alex Cloud, Alex Turner, and Jacob Goldman-Wetzler for discussions on gradient routing. We are grateful to Ryan Kidd and the ICML Technical AI Governance Workshop for funding support that enabled the presentation of this work.

\bibliography{references}
\bibliographystyle{packages/icml2025}

\newpage
\appendix
\crefalias{section}{appendix}

\section{Estimating the Number of Requests Related to the Biology of Pathogens} \label{appendix:estimating-biology-requests}

To estimate how many user requests are related to the biology of pathogens, we used the second version of the Anthropic Economic Index \cite{handa2025economictasksperformedai}, a dataset of 1 million anonymized conversations from the Free and Pro tiers of Claude.ai.
In the dataset, the conversations are clustered by topic, and the proportion of each topic in the whole dataset is given.
For example, the topic ``Help with agricultural business, research, and technology projects'' makes up 0.15\% of the requests in the dataset.
There are three levels of topic granularity; we use the lowest, most granular level.

We filtered the dataset to only include conversations whose topic contains one of the following keywords related to biology: \emph{cell} (when at the beginning of the word), \emph{genet}, \emph{genom}, \emph{microb}, \emph{bacteria}, \emph{virus}, \emph{viral}, \emph{proteo}, \emph{protei}, \emph{immune}, \emph{neuro}, \emph{patho}, \emph{infect}; we also required that it does not contain any of the following keywords to avoid false positives: \emph{nutri}, \emph{tweet}, \emph{agric}, \emph{sexual health}.
The total proportion of these requests was 0.85\%.
When we applied similar methodology to identify requests related to any kind of biology, the proportion was 2.98\%.

\end{document}